%% file: 0.main.tex
\documentclass[10pt,twocolumn,letterpaper]{article}

\usepackage{iccv}
\input{configs}

\iccvfinalcopy 

\def\iccvPaperID{5029} 

\ificcvfinal\fi

\begin{document}

\title{3D-VisTA: Pre-trained Transformer for 3D Vision and Text Alignment}

\author{
\begin{tabular}{cccccc}
Ziyu Zhu$^{1}$\thanks{Work done as an intern at BIGAI.~~~\Letter~Corresponding author.} & Xiaojian Ma$^{2}$ & Yixin Chen$^{2}$ & Zhidong Deng$^{1\text{\Letter}}$ & Siyuan Huang$^{2\text{\Letter}}$ & Qing Li$^{2\text{\Letter}}$
\end{tabular}
\\
\begin{tabular}{cc}
$^1$Tsinghua University & $^2$National Key Laboratory of General Artificial Intelligence, BIGAI, China
\end{tabular}
\\
\href{https://3d-vista.github.io}{3d-vista.github.io}
}

\maketitle
\ificcvfinal\thispagestyle{empty}\fi

\begin{abstract}
3D vision-language grounding (3D-VL) is an emerging field that aims to connect the 3D physical world with natural language, which is crucial for achieving embodied intelligence. Current 3D-VL models rely heavily on sophisticated modules, auxiliary losses, and optimization tricks, which calls for a simple and unified model. 
In this paper, we propose \model, a pre-trained Transformer for \underline{3D} \underline{Vis}ion and \underline{T}ext \underline{A}lignment that can be easily adapted to various downstream tasks. \model simply utilizes self-attention layers for both single-modal modeling and multi-modal fusion without any sophisticated task-specific design. 
To further enhance its performance on 3D-VL tasks, we construct ScanScribe, the first large-scale 3D scene-text pairs dataset for 3D-VL pre-training. ScanScribe contains 2,995 RGB-D scans for 1,185 unique indoor scenes originating from ScanNet and 3R-Scan datasets, along with paired 278K scene descriptions generated from existing 3D-VL tasks, templates, and GPT-3.
\model is pre-trained on ScanScribe via masked language/object modeling and scene-text matching. It achieves state-of-the-art results on various 3D-VL tasks, ranging from visual grounding and dense captioning to question answering and situated reasoning. Moreover, 3D-VisTA demonstrates superior data efficiency, obtaining strong performance even with limited annotations during downstream task fine-tuning.
\end{abstract}

\input{1.introduction}
\input{2.related-work}
\input{3.method}
\input{4.experiments}
\input{5.conclusions}

\clearpage
\noindent\textbf{Acknowledgements.}
The authors would like to thank Hongming Xu at BIGAI for the help on Mask3D. This work is supported in part by the National Key R\&D Program of China (2022ZD0114900) and the National Science Foundation of China (NSFC) under Grant No. 62176134.

{\small
\bibliographystyle{ieee_fullname}
\bibliography{reference}
}

\input{appendix}

\end{document}

%% file: configs.tex
\usepackage[dvipsnames]{xcolor}
\definecolor{Gray}{gray}{0.9}
\definecolor{mygreen}{rgb}{0.0, 0.5, 0.0}
\definecolor{myred}{rgb}{0.8, 0.25, 0.33}
\definecolor{myblue}{rgb}{0.19, 0.55, 0.91}
\definecolor{uclablue}{rgb}{0.15, 0.45, 0.68}
\definecolor{boxgreen}{rgb}{0.02, 0.66, 0.02}
\definecolor{boxred}{rgb}{0.66, 0.1, 0.1}
\definecolor{boxblue}{rgb}{0.01, 0.01, 0.73}
\definecolor{ballblue}{rgb}{0.13, 0.67, 0.8}

\usepackage{times}
\usepackage{microtype}
\usepackage{graphicx} \graphicspath{{figures/}}
\usepackage{amsmath,amssymb,amsbsy,mathtools,etoolbox}
\usepackage{algorithm,algorithmic}
\usepackage{acronym}
\usepackage{enumitem}
\usepackage[pagebackref,breaklinks,citecolor=uclablue,colorlinks=true,linkcolor=blue,bookmarks=false]{hyperref}
\usepackage{balance}
\usepackage{xspace}
\usepackage{setspace}
\usepackage[skip=3pt,font=small]{subcaption}
\usepackage[skip=3pt,font=small]{caption}
\usepackage[capitalise,nameinlink]{cleveref}
\usepackage{booktabs,tabularx,colortbl,multirow,multicol,array,makecell}
\usepackage{overpic,wrapfig}
\usepackage{dblfloatfix}
\usepackage[misc]{ifsym}
\usepackage{epsfig}
\usepackage{array}
\usepackage{booktabs}
\usepackage{bbding}
\usepackage{pifont}
\usepackage{subfiles}

\makeatletter
\newcommand*{\addFileDependency}[1]{
  \typeout{(#1)}
  \@addtofilelist{#1}
  \IfFileExists{#1}{}{\typeout{No file #1.}}
}
\makeatother

\makeatletter
\DeclareRobustCommand\onedot{\futurelet\@let@token\@onedot}
\def\@onedot{\ifx\@let@token.\else.\null\fi\xspace}
\def\eg{\emph{e.g}\onedot} 

\def\ie{\emph{i.e}\onedot}

\def\wrt{w.r.t\onedot} 

\def\etal{\emph{et al}\onedot}
 
\makeatother

\makeatletter
\renewcommand{\paragraph}{%
  \@startsection{paragraph}{4}%
  {\z@}{0ex \@plus 0ex \@minus 0ex}{-1em}%
  {\hskip\parindent\normalfont\normalsize\bfseries}%
}
\makeatother

\usepackage{calc}
\newlength\myheight
\newlength\mydepth
\settototalheight\myheight{Xygp}
\newcommand{\model}{3D-VisTA\xspace}

\newcommand{\inc}[1]{\textcolor{blue}{\textbf{#1 $\uparrow$}}}
\newcommand{\dec}[1]{\textcolor{red}{\textbf{#1 $\downarrow$}}}

%% file: 1.introduction.tex
\section{Introduction}
\begin{figure}[!t]
    \centering
    \includegraphics[width=\linewidth]{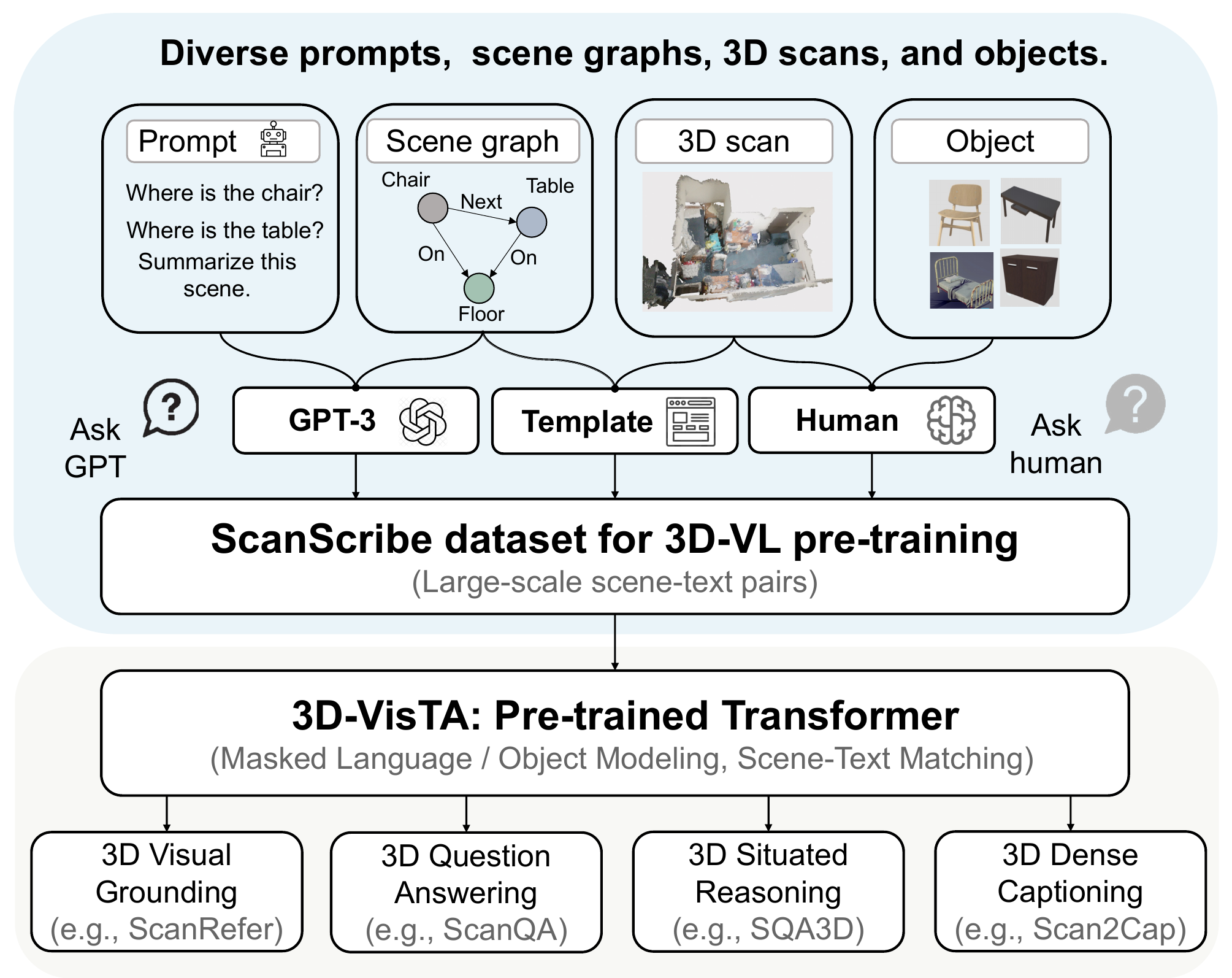}
    \caption{Overall framework of our \model pipeline. We collect diverse prompts, scene graphs, 3D scans, and objects to construct ScanScribe dataset. Through self-supervised pre-training, \model supports various downstream tasks including 3D visual grounding, dense captioning, question answering, and situated reasoning.} 
    \label{fig:overall}
\end{figure}

Aligning the 3D physical world with natural language is a crucial step towards embodied artificial intelligence~\cite{embodied, embodiedlang, teach}, where intelligent agents can understand and further execute human instructions in the real world~\cite{saycan, vima}. Recently, 3D vision-language (3D-VL) tasks have attracted growing interest~\cite{semab}, including 3D visual grounding~\cite{scanrefer, referit3d}, dense captioning~\cite{scan2cap}, grammar learning~\cite{hong2021vlgrammar}, question answering~\cite{scanqa, 3dgqa}, and situated reasoning~\cite{sqa3d}.

However, most of the models developed for 3D-VL only focus on one or two of these 3D-VL tasks and employ task-specific designs ~\cite{3djcg, scanqa, sqa3d, 3dsps, vil3dref}. For instance, 3D-SPS~\cite{3dsps} and BUTD-DETR~\cite{butd} progressively discover the target object by attending VL features and detecting objects in each layer. 3DVG~\cite{3dvg}, MVT~\cite{mvt3d}, and ViL3DRel~\cite{vil3dref} improve 3D visual grounding by explicitly infusing spatial relation information into the model design.
3DJCG~\cite{3djcg} jointly learns 3D dense captioning and visual grounding via a shared 3D object proposal module~\cite{votenet} with two separate task-specific heads~\cite{3djcg}. 
Additionally, training these models often requires manually specified auxiliary losses (\eg, 3D object detection/classification and text classification~\cite{3dsps, mvt3d, 3djcg, scanqa, sqa3d}) or optimization tricks (\eg, knowledge distillation~\cite{lar, sat} ). The lack of a simple and unified approach creates a significant gap in developing a general-purpose 3D-VL model. 

To fill such gap, we introduce \textbf{\model}, a Transformer-based model for \underline{3D} \underline{Vis}ion and \underline{T}ext \underline{A}lignment that can be easily adapted to various downstream tasks. Unlike previous models that design sophisticated task-specific modules, we simply utilize a vanilla self-attention transformer \cite{vaswani2017attention} for both single-modal modeling and multi-modal fusion in the \model. As a general approach to further enhance 3D spatial comprehension \cite{vil3dref, 3dvg, 3djcg}, we explicitly encode the pairwise spatial relations between objects into the self-attention weights for 3D object modeling.

Inspired by the success of large-scale pre-training in NLP~\cite{bert,gpt-1,gpt-2,gpt-3,yang2019xlnet,liu2019roberta}, CV~\cite{he2020momentum,dosovitskiy2021image,mae,huang2021spatio,pointmae}, and 2D-VL~\cite{oscar,flamingo,lu2019vilbert,radford2021learning}, we propose to pre-train \model on 3D scene-text data, aiming for better performances on 3D-VL tasks. To this end, we construct ScanScribe, the first large-scale 3D scene-text pairs dataset for 3D-VL pre-training. We first collect RGB-D scans of indoor scenes from ScanNet~\cite{dai2017scannet} and 3R-Scan~\cite{wald2019rio} datasets. We also randomly replace some objects in the scene with objects from the Objaverse 3D object database~\cite{deitke2022objaverse} based on their categories, in order to increase object diversity. To obtain the text, we transform the text from existing datasets based on ScanNet into scene descriptions, including the question-answer pairs from ScanQA~\cite{scanqa} and the referring expressions from ScanRefer~\cite{scanrefer} and ReferIt3D~\cite{referit3d}. We further leverage the scene graph annotations \cite{scenegraph} of scans from 3R-Scan, and adopt both templates and GPT-3~\cite{gpt-3} to generate scene descriptions from their scene graphs. In total, ScanScribe contains 278K 3D scene-text pairs for 2,995 RGB-D scans of 1,185 indoor scenes, with 56.1K unique object instances.

We pre-train \model on the proposed ScanScribe dataset. Our pre-training tasks include masked language modeling, masked object modeling, and scene-text matching.  Notably, similar objectives are widely adopted in 2D-VL yet rarely explored in the 3D-VL domain. The proposed pre-training procedure effectively learns the alignment between 3D point clouds and texts, which eliminates the need for auxiliary losses and optimization tricks in downstream task fine-tuning. On six challenging 3D-VL tasks, ranging from visual grounding (\ie, ScanRefer \cite{scanrefer}, Nr3D/Sr3D \cite{referit3d}) and dense captioning (\ie, Scan2Cap~\cite{scan2cap}) to question answering (\ie, ScanQA \cite{scanqa}) and situated reasoning (\ie, SQA3D \cite{sqa3d}), fine-tuned \model raises the SOTA results on ScanRefer by 8.1\% (acc@0.5), on Sr3D by 3.6\%, on Scan2Cap by 10.1\%(C@0.25), on ScanQA by 3.5\%/2.1\% (EM@1), and on SQA3D by 1.9\%. Moreover, 3D-VisTA demonstrates superior data efficiency, obtaining strong results with only 30\% of the annotations for these downstream tasks.

Our main contributions can be summarized as follows:
\begin{itemize}[leftmargin=*,noitemsep,topsep=0pt]
    \item We propose \model, a simple and unified Transformer for aligning 3D vision and text. The proposed Transformer simply utilizes the self-attention mechanism, without any complex task-specific design. 
    \item We construct ScanScribe, a large-scale 3D-VL pre-training dataset that contains 278K 3D scene-text pairs for 2,995 RGB-D scans of 1,185 unique indoor scenes.
    \item We introduce a self-supervised pre-training scheme for 3D-VL, with masked language/object modeling and scene-text matching. It effectively learns the 3D point cloud and text alignment and further simplifies and improves downstream task fine-tuning.
    \item We fine-tune \model and achieve state-of-the-art performances on various 3D-VL tasks, ranging from visual grounding and dense captioning to question answering and situated reasoning. \model also demonstrates superior data efficiency, obtaining strong results even with limited annotations.
\end{itemize}

%% file: 2.related-work.tex
\section{Related Work}
\begin{table*}[!t]
\begin{minipage}[t]{0.65\textwidth}
\centering
\small
\caption{The comparison between \model and other models \wrt tasks, auxiliary losses, and task-specific architectures.``VG'' stands for visual grounding, ``QA'' for question answering, ``SR'' for situation reasoning, ``DC'' for dense captioning. ``DET'' stands for object detection loss, ``KD'' for knowledge distillation loss, ``O-CLS'' for object classification loss, and ``T-CLS'' for text classification loss. ``CA'' stands for cross attention, ``2D'' for 2D features, ``MV'' for multi-view features, and ``LC'' for language-conditioned modules.} \label{tab:model_comparison}
\resizebox{\linewidth}{!}{
\begin{tabular}{c|c|cccc|cccc}
\toprule
\multirow{2}{*}{\textbf{Method}} & \multirow{2}{*}{\textbf{Task}} & \multicolumn{4}{c|}{\textbf{Auxiliary loss}} & \multicolumn{4}{c}{\textbf{Architecture}} \\
 &  & DET & KD & O-CLS & T-CLS & CA & 2D & MV & LC \\ \midrule
MVT~\cite{mvt3d} & VG &  &  & \checkmark & \checkmark & \checkmark &  & \checkmark &  \\
3D JCG~\cite{3djcg} & VG, DC & \checkmark &  & \checkmark & \checkmark & \checkmark & \checkmark &  &  \\
ViL3DRel~\cite{vil3dref} & VG &  & \checkmark & \checkmark & \checkmark & \checkmark &  &  & \checkmark \\
ScanQA~\cite{scanqa} & QA & \checkmark &  & \checkmark &  & \checkmark & \checkmark &  &  \\
SQA3D~\cite{sqa3d} & SR & \checkmark &  &  &  & \checkmark &  &  &  \\ \hline
3D-VisTA (ours) & VG,QA,SR,DC & $\times$ & $\times$ & $\times$ & $\times$ & $\times$ & $\times$ & $\times$ & $\times$ \\ \bottomrule
\end{tabular}
}
\end{minipage}
\hfill
\begin{minipage}[t]{0.33\textwidth}
\centering
\small
\caption{The comparison between ScanScribe and other 3D-VL datasets. ``VG'' stands for Visual Grounding, ``QA'' for Question Answering, ``SR'' for Situated Reasoning, and ``PT'' for Pre-training. ``Vocab.'' denotes the text vocabulary size.} \label{tab_dataset_comp}
\resizebox{\linewidth}{!}{
\begin{tabular}{cccc}
\toprule
\textbf{Dataset} & \textbf{Task} & \textbf{Size} & \textbf{Vocab.} \\
\midrule
Nr3D~\cite{referit3d} & VG & 30.0K & 2,986 \\
Sr3D~\cite{referit3d}& VG & 90.5K & 158 \\
ScanRefer~\cite{scanrefer} & VG & 36.7K & 4,197 \\
ScanQA~\cite{scanqa} & QA & 26.5K & 3,357  \\
SQA3D~\cite{sqa3d} & SR & 33.4K & 4,535 \\
\midrule
ScanScribe & PT & 278.0K & 8,197 \\
\bottomrule
\end{tabular}

}
\end{minipage}
\end{table*}

\noindent\textbf{3D Vision-language Learning.}
Recently, there has been growing interest in 3D vision-language (3D-VL) learning. Unlike traditional scene understanding, 3D-VL tasks connect the physical world to natural language, which is crucial for achieving embodied intelligence~\cite{embodied}.
In this emerging area, Chen \etal~\cite{scanrefer} and Achlioptas \etal~\cite{referit3d} concurrently introduce ScanRefer and ReferIt3D datasets for benchmarking natural language grounding to 3D object properties and relations. 
Besides 3D visual grounding, Azuma \etal~\cite{scanqa} develop a 3D question-answering dataset named ScanQA that requires a model to answer a question about objects and their relations given a 3D scene. More recently, Ma \etal~\cite{sqa3d} propose a situated reasoning task called SQA3D for embodied scene understanding in 3D scenes.

Several models have been proposed for these benchmarks~\cite{scanrefer, referit3d, 3dsps, butd, 3dvg, mvt3d, vil3dref, transrefer3d, roh2022languagerefer}. Notably, 3D-SPS~\cite{3dsps} and BUTD-DETR~\cite{butd} progressively discover the target object by leveraging cross attention mechanism and language guidance. 3DVG~\cite{3dvg}, MVT~\cite{mvt3d}, and ViL3DRel~\cite{vil3dref} tackle 3D visual grounding by explicitly infusing spatial relation information into their models.
Although these works have achieved impressive results in bridging 3D vision and language, they still rely heavily on task-specific knowledge in model design \cite{3dvg, mvt3d, vil3dref} and sophisticated optimization techniques \cite{vil3dref, butd, 3dsps}. In contrast, the proposed \model unifies visual grounding, question-answering, and situated reasoning through a simple Transformer-based architecture. Training \model is also straightforward, without requiring any auxiliary losses or sophisticated optimization techniques. Refer to \cref{tab:model_comparison} for a detailed comparison between \model and other 3D-VL models \wrt task, auxiliary Loss, and architecture.

\noindent \textbf{Large-scale Pre-training.}
In recent years, large-scale pre-training has become a cornerstone of natural language processing (NLP), computer vision (CV), and 2D vision-and-language (2D-VL) domains. The introduction of the transformer-based architecture~\cite{transformer}, especially BERT~\cite{bert} and GPT~\cite{gpt-1,gpt-2,gpt-3}, has led to significant improvements in various NLP tasks. The success of these models has led to the development of more advanced pre-training techniques such as XLNet~\cite{yang2019xlnet} and RoBERTa~\cite{liu2019roberta}. These models have achieved state-of-the-art performance on a wide range of NLP tasks, including text classification, question answering, and language generation. The most successful pre-training approach in CV is the ImageNet~\cite{imagenet} pre-training, which has been used as a starting point for a wide range of downstream tasks such as object detection and image segmentation. Recently, the introduction of transformer-based models such as ViT~\cite{dosovitskiy2021image} and Swin Transformer~\cite{liu2021swin} has led to significant improvements in various CV tasks.
The field of 2D-VL has also seen significant advancements due to pre-training techniques. In particular, the introduction of the ViLBERT~\cite{lu2019vilbert} and LXMERT~\cite{tan2019lxmert} models has led to state-of-the-art performance on tasks such as visual question answering and image captioning. More recently, the development of CLIP~\cite{radford2021learning}, ALIGN~\cite{wang2022align}, and Flamingo~\cite{flamingo} has shown that large-scale pre-training on image-text pairs leads to better cross-modal understanding and the emerge of in-context learning in a zero-shot or few-shot manner.  

Although large-scale pre-training has become a crucial technique in NLP, CV, and 2D-VL, it has rarely been explored in 3D-VL. \cite{3djcg, d3net} explore multi-task learning of visual grounding and dense captioning, and then further fine-tune their models on each task. The exploration of 3D-VL pre-training may be hindered by the lack of a large-scale pre-training dataset. Therefore, we construct ScanScribe, the first large-scale 3D scene-text pairs dataset for 3D-VL pre-training. As shown in \cref{tab_dataset_comp}, ScanScribe is much larger than existing 3D-VL datasets and also has more diverse text. Pre-training \model on ScanScribe has led to significant improvements on 3D-VL tasks, so we believe ScanScribe can fuel the exploration of 3D-VL pre-training in the future.

%% file: 3.method.tex
\section{\model}
\begin{figure*}[!ht]
    \centering
    \includegraphics[width=\linewidth]{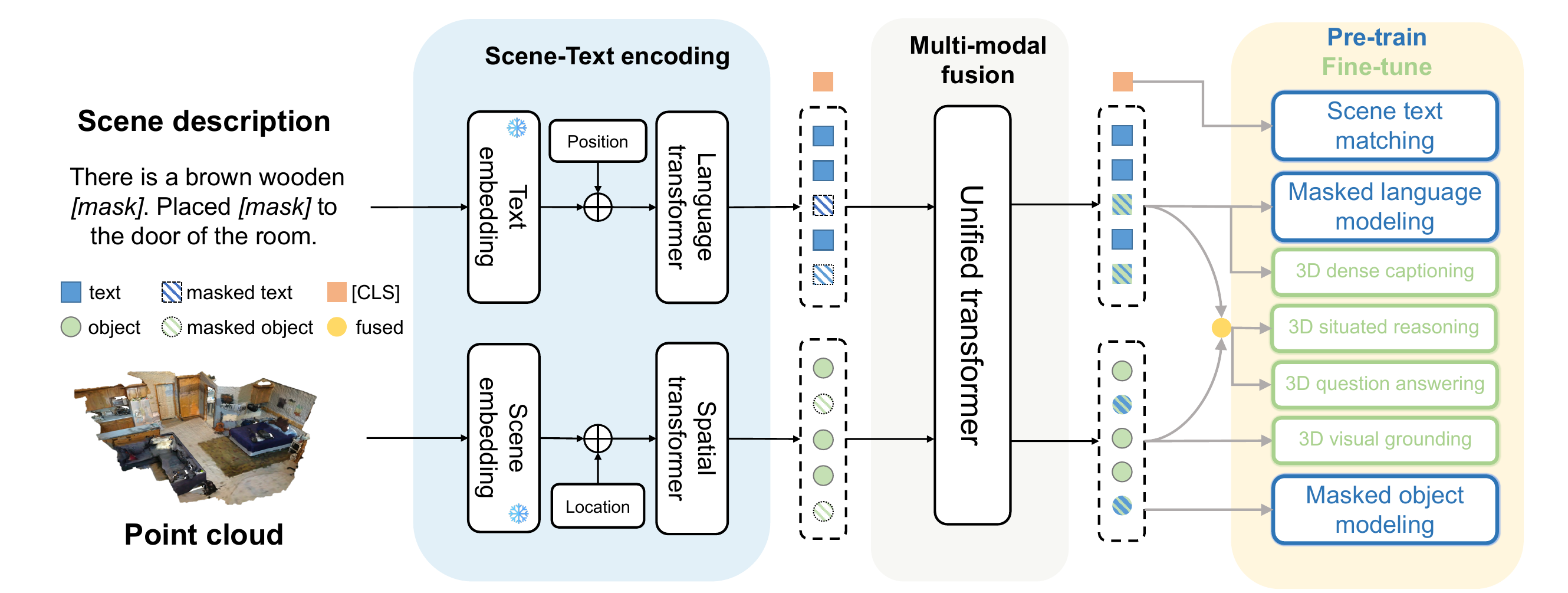}
    \caption{The model architecture of our \model, which includes text encoding, scene encoding, and multi-modal fusion modules. \model is pre-trained by self-supervised learning objectives, which include masked language modeling, masked object modeling, and scene-text matching. Pre-trained \model can be easily adapted to various downstream tasks by adding lightweight task heads without task-specific design like auxiliary losses and optimization tricks.}
    \label{fig:model}
\end{figure*}

In this section, we introduce \model, a simple and unified Transformer for aligning 3D scenes and text. As illustrated by \cref{fig:model}, \model takes a pair of scene point cloud and sentence as input. It first encodes the sentence via a text encoding module and processes the point cloud via a scene encoding module. Then the text and 3D object tokens are fused by a multi-modal fusion module to capture the correspondence between 3D objects and text. \model is pre-trained using self-supervised learning and can be easily fine-tuned to various downstream tasks. Next, we describe each module in detail.

\subsection{Text Encoding}
We adopt a four-layer Transformer to encode the sentence $S$ into a sequence of text tokens $\{w_{\text{cls}}, w_1, w_2, \cdot \cdot \cdot, w_M \}$, where $w_{cls}$ is a special classification token (\texttt{[CLS]}) and $M$ is the sentence length. This text encoding module is initialized by the first four layers of a pre-trained BERT~\cite{bert}.

\subsection{Scene Encoding}
Given the point cloud of a 3D scene, we first use segmentation masks to break down the scene into a bag of objects. The segmentation masks can be either obtained from ground truth or instance segmentation models \cite{votenet,pointgroup,mask3d}. For each object, we sample 1024 points and normalize their coordinates into a unit ball. Then the object point cloud is fed into PointNet++ \cite{pointnet++} to obtain its point features and semantic class. We compose the point features $f_i$, the semantic class embedding $c_i$, and the location $l_i$ (\ie, 3D position, length,  width, height) as the representation of the object token $i$: 
\begin{equation} \label{object_token}
o_i = f_i + W_cc_i  + W_l l_i, i = 1, 2, ..., N,
\end{equation}
where $W_c$ and $W_l$ are additional projection matrices to map $c_i$ and $l_i$ into the same dimension as $f_i$.

To further provide a contextual representation of objects, we capture the object-to-object interactions by infusing object tokens into a four-layer Transformer. Motivated by previous works \cite{3dvg,mvt3d,vil3dref}, we explicitly encode the pairwise spatial relations of objects into the Transformer (\textit{Spatial transformer} in \cref{fig:model}). More specifically, we follow \cite{vil3dref} to define the pairwise spatial features for the object pair $i, j$:
\begin{align*}
    s_{ij} = [d_{i j}, \sin (\theta_{h}), \cos (\theta_{h}), \sin (\theta_{v}), \cos (\theta_{v})],
\end{align*}
where $d_{i j}$ is the Euclidean distance and $\theta_{h}, \theta_{v}$ are the horizontal and vertical angles of the line connecting the centers of objects $i,j$. The pairwise spatial features $S =[s_{ij}] \in \mathbb{R}^{N \times N \times 5}$ are used to modulate the attention weights of the self-attention layers in the Transformer:
\begin{align*}
\operatorname{Attn}(Q,K,V,S)=\operatorname{softmax}\left(\frac{Q K^{T}}{\sqrt{d_{h}}} + \log\sigma(Sw)\right) V,
\end{align*}
where $w \in \mathbb{R}^5$ is used to map the spatial features to the attention scores and $\sigma$ is the sigmoid function.

\subsection{Multi-modal Fusion}
We simply concatenate the text and the 3D object tokens and send them to a $L$-layer Transformer (\textit{Unified transformer} in \cref{fig:model}) for multi-modal fusion. Learnable type embeddings are added to the tokens to differentiate text and 3D objects. We denote the output of the multi-modal fusion module as $\{\textbf{w}_{\text{cls}}, \textbf{w}_{1:M}, \textbf{o}_{1:N}\}$ for \texttt{[CLS]}, text tokens, and 3D object tokens, respectively.

\subsection{Self-supervised Pre-training}
To learn the 3D scene and text alignment in a self-supervised manner, we pre-train \model on 3D scene-text pairs via the following proxy tasks:

\noindent
\textbf{Masked Language Modeling (MLM).} We follow the BERT pre-training \cite{bert} to perform MLM: (1) 15\% of the text tokens are randomly chosen; (2) 80\% of the time: replace these tokens with \texttt{[MASK]}; (2) 10\% of the time: replace these tokens with some random text tokens; (3) 10\% of the time: these tokens remain unchanged. The model is trained to predict the masked text tokens given the remaining text and 3D object tokens:

\begin{equation}
\mathcal{L}_{\mathrm{MLM}}=-\mathbb{E}_{(\mathbf{w}, \mathbf{o}) \sim D} \log P_\theta\left(\mathbf{w}_{\mathbf{m}} \mid \mathbf{w}_{\backslash \mathbf{m}}, \mathbf{o}\right).
\end{equation}

\noindent\textbf{Masked Object Modeling (MOM).} Similar to MLM, we mask out 10\% of 3D object tokens. However, we mask a 3D object token by only replacing its point features and semantic embedding (\ie, ``$f_i + W_cc_i$'' in \cref{object_token}) with a learnable mask embedding but keep its positional information (\ie, ``$W_ll_i$'' in \cref{object_token}) unchanged. The model is trained to utilize the position clue of the masked object to predict its semantic class $c$ given the remaining 3D objects and text:

\begin{equation}
\mathcal{L}_{\mathrm{MOM}}=-\mathbb{E}_{(\mathbf{w}, \mathbf{o}) \sim D} \log P_\theta\left(c(\mathbf{o}_{\mathbf{m}}) \mid \mathbf{o}_{\backslash \mathbf{m}}, \mathbf{w}\right).
\end{equation}

\noindent\textbf{Scene-Text Matching (STM).}  While masked language and object modeling enable local text-object alignment in a fine-grained granularity, we also perform scene-text matching to enhance the global fusion of scene and text, which we find very beneficial for downstream question-answering tasks. More specifically, we extract the output corresponds to \texttt{[CLS]} as the global representation of the input scene-text pair, and feed it into a two-layer MLP to predict if the scene and the text are matched:

\begin{equation}
\mathcal{L}_{\mathrm{STM}}=-\mathbb{E}_{(\mathbf{w}, \mathbf{o}) \sim D} \log P_\theta\left(y \mid \mathbf{w}, \mathbf{o}\right).
\end{equation}
In practice, 30\% of the samples in a training batch are negative pairs, created by replacing the scene point cloud or text with a randomly selected sample.

\noindent \textbf{Final loss.}  Our final pre-training objective is obtained by simply adding the losses of the proxy tasks above: 
\begin{equation}
    \mathcal{L}_\text{pre-train} = \mathcal{L}_\text{MLM} + \mathcal{L}_\text{MOM} + \mathcal{L}_\text{STM}
\end{equation}
Notably, the proposed pre-training scheme is self-supervised and task-agnostic, unlike the supervised multi-task learning used in previous work~\cite{3djcg} that requires task supervision.

\subsection{Downstream Task Finetuning}
The pre-trained \model can be easily adapted to various 3D-VL tasks by adding lightweight task heads. More specifically, we fine-tune \model on the following tasks:

\noindent\textbf{3D Visual Grounding} tasks a model to locate a target object in a 3D scene from a referring expression. To find the referred object, we apply a two-layer MLP to each object token $\textbf{o}_{i}$, and obtain the probability of the object being referred to.  The model is fine-tuned using the cross-entropy loss.

\noindent\textbf{3D Dense Captioning} is introduced by~\cite{scan2cap} to test a model's ability of detecting and describing objects in a 3D scene. Following \cite{oscar}, we take $\mathbf{w}_{1: M}$ and predict text tokens autoregressively to generate a sentence. The model is fine-tuned using cross-entropy loss.

\noindent\textbf{3D Question Answering} requires a model to answer an object-related question given a 3D scene. Following~\cite{scanqa}, we feed the text tokens $\textbf{w}_{1:M}$ and the object tokens $\textbf{o}_{1:N}$ into a modular co-attention network (MCAN)~\cite{mcan} to produce answers. The model is fine-tuned using the QA loss and the object localization loss.

\noindent\textbf{3D Situated Reasoning} is recently proposed by~\cite{sqa3d} to benchmark the 3D scene understanding of embodied agents. 
To adapt \model to this task, we concatenate the situation description and the question into a single input sentence. The answer classification is similar to the 3D question answering task. The model is fine-tuned using the answer loss.

In general, we find adapting \model to these downstream tasks much simpler than previous methods~\cite{scanrefer,mvt3d,vil3dref,scanqa,sqa3d}, as \model is simply fine-tuned using the task loss only, without the need for any auxiliary losses (\eg, sentence/object classification loss~\cite{scanrefer,scanqa}) or optimization tricks (\eg, multi-view aggregation~\cite{mvt3d} and knowledge distillation~\cite{vil3dref}). This makes \model a more unified and general-purpose 3D-VL model.

\begin{table}[t!]
\centering
\small
\caption{The composition of ScanScribe. $^*$We only use Objaverse to provide candidate object replacement for the 3D scenes in other two datasets; thus no scene-text pair is generated.} \label{tab_construction}
\resizebox{\linewidth}{!}{%
\setlength\tabcolsep{2pt}
\begin{tabular}{l|ccc|ccc|c}
\toprule
\multirow{2}{*}{\textbf{Source}} & \multicolumn{3}{c|}{\textbf{3D}} & \multicolumn{3}{c|}{\textbf{Text}} & \textbf{Scene-Text} \\
 & \textbf{Scan} & \textbf{Scene} & \textbf{Object} & \textbf{Human} & \textbf{Template} & \textbf{GPT-3} & \textbf{Pairs} \\ 
 \midrule
ScanNet & 1,513 & 707  & 36.2K & 93.2K & 90.5K & - & 183.7K  \\
3R-Scan & 1,482 & 478 & 13.6K & - & 89.6K & 4.7K & 94.3K \\ 
Objaverse$^*$ & - & - & 6.3K & - & - & - & - \\
\midrule
ScanScribe & 2,995  & 1,185  & 56.1K & 93.2K & 180.1K & 4.7K  & 278.0K \\ 
\bottomrule
\end{tabular}
}
\end{table}

\section{ScanScribe} \label{sec:scancribe}
In recent years, large-scale pre-training has been widely used to improve the performance on downstream tasks in CV~\cite{beit}, NLP~\cite{bert}, and 2D-VL~\cite{oscar,tan2019lxmert}. However, large-scale pre-training has barely been touched in the 3D-VL domain, possibly due to the lack of pre-training datasets for 3D-VL. To facilitate the exploration of 3D-VL pre-training, we build a large-scale 3D scene-text pairs dataset, named ScanScribe. As illustrated in \cref{tab_construction}, the construction of 3D scene-text pairs in ScanScribe comprises two parts:

\noindent\textbf{3D scenes.} We collect RGB-D scans of indoor scenes from ScanNet \cite{dai2017scannet} and 3R-Scan \cite{wald2019rio}. To increase the diversity of 3D objects in these scenes, 10\% of the object instances in each scene are randomly replaced by objects from the Objaverse 3D object database\cite{deitke2022objaverse} based on their categories. For each ScanNet and 3R-Scan object category, we download about 40 object instances from Objaverse as candidate object replacements. As a result, we collect 2,995 RGB-D scans of 1,185 indoor scenes, with 56.1K unique object instances.

\noindent\textbf{Text.} For the scans from ScanNet, we transform the text from existing datasets based on ScanNet into scene descriptions, including the question-answer pairs from ScanQA \cite{scanqa} and the referring expressions from ScanRefer \cite{scanrefer} and ReferIt3D~\cite{referit3d}. For the scans from 3R-Scan, we adopt both templates and GPT-3~\cite{gpt-3} to generate scene descriptions based on their scene graph annotations \cite{scenegraph}. Specifically, for each \texttt{object}, we first extract all the $\langle\text{\texttt{object}, \texttt{relation}, \texttt{neighbor}}\rangle$ triplets from the scene graph. We then use the template ``This is a \texttt{object}, a \texttt{neighbor} is \texttt{relation} to \texttt{object}'' to generate the descriptions. Note that we only choose objects with fewer than 7 neighbors in a template-based generation. We further explore using GPT-3 to generate the descriptions with the following prompt ``\texttt{object} is \texttt{relation} to \texttt{neighbor} ...\textit{(repeat until all the neighbors have been used)}. Where is \texttt{object}? or Summarize the scene.'' Ultimately, 278K scene descriptions are generated for the collected 3D scenes. 


%% file: 4.experiments.tex
\begin{table*}[!ht]
    \centering
    \small
    \caption{Grounding accuracy (\%) on Nr3D and Sr3D with ground-truth object proposals. $\Delta$ denotes the performance difference between \model and \model (scratch). \model achieves competitive results with SOTA on Nr3D and outperforms SOTA on Sr3D. } \label{tab:nrsr}
    \begin{tabular}{lccccc|ccccc}
        \toprule
        \multirow{3}{*}{Method} & \multicolumn{5}{c|}{Nr3D} & \multicolumn{5}{c}{Sr3D} \\ \cline{2-11} 
        & \multirow{2}{*}{Overall} & \multirow{2}{*}{Easy} & \multirow{2}{*}{Hard} & \multirow{2}{*}{\begin{tabular}[c]{@{}c@{}}View\\ Dep\end{tabular}} & \multirow{2}{*}{\begin{tabular}[c]{@{}c@{}}View\\ Indep\end{tabular}} 
        & \multirow{2}{*}{Overall} & \multirow{2}{*}{Easy} & \multirow{2}{*}{Hard} & \multirow{2}{*}{\begin{tabular}[c]{@{}c@{}}View\\ Dep\end{tabular}} & \multirow{2}{*}{\begin{tabular}[c]{@{}c@{}}View\\ Indep\end{tabular}} \\
        & & & & & & & & & & \\
        \midrule
        3DVG-Trans \cite{3dvg} & 40.8 & 48.5 & 34.8 & 34.8 & 43.7 & 51.4 & 54.2 & 44.9 & 44.6 & 51.7 \\
        TransRefer3D \cite{transrefer3d} & 48.0 & 56.7 & 39.6 & 42.5 & 50.7 & 57.4 & 60.5 & 50.2 & 49.9 & 57.7 \\
        LAR \cite{lar} & 48.9 & 58.4 & 42.3 & 47.4 & 52.1 & 59.4 & 63.0 & 51.2 & 50.0 & 59.1 \\
        SAT \cite{sat} & 56.5 & 64.9 & 48.4 & 54.4 & 57.6 & 57.9 & 61.2 & 50.0 & 49.2 & 58.3 \\
        3D-SPS \cite{3dsps} & 51.5 & 58.1 & 45.1 & 48.0 & 53.2 & 62.6 & 56.2 & 65.4 & 49.2 & 63.2 \\
        MVT \cite{mvt3d} & 59.5 & 67.4 & 52.7 & 59.1 & 60.3 & 64.5 & 66.9 & 58.8 & 58.4 & 64.7 \\
        ViL3DRel \cite{vil3dref} & \textbf{64.4} & 70.2 & \textbf{57.4} & \textbf{62.0} & 64.5 & 72.8 & 74.9 & 67.9 & \textbf{63.8} & 73.2 \\ 
        \midrule
        \model (scratch) & 57.5 & 65.9 & 49.4 & 53.7 & 59.4 & 69.6 & 72.1 & 63.6 & 57.9 & 70.1 \\
        \model & 64.2 & \textbf{72.1} & 56.7 & 61.5 & \textbf{65.1} & \textbf{76.4} & \textbf{78.8} & \textbf{71.3} & 58.9 & \textbf{77.3} \\ 
        $\Delta$ & \inc{6.7} & \inc{6.2} & \inc{7.3} & \inc{7.8} & \inc{5.7} & \inc{6.8} & \inc{6.7} & \inc{7.7} & \inc{1.0} & \inc{7.2} \\
        \bottomrule
    \end{tabular}
\end{table*}

\begin{table*}[!ht]
    \centering
    \small
    \caption{Grounding accuracy (\%) on ScanRefer with detected object proposals. ``Det.'' represents the 3D object detection module used in the model. ``VN'' stands for VoteNet~\cite{votenet}, ``PG'' for PointGroup~\cite{pointgroup}, and M3D for Mask3D~\cite{mask3d}, while ``Opt.'' denotes jointly optimizing the object detector on ScanRefer. Mask3D significantly improves the grounding accuracy by providing more accurate object proposals.}\label{tab:scanrefer}
    \begin{tabular}{lccccccc}
        \toprule
        \multirow{2}{*}{Method} & \multirow{2}{*}{Det.} & \multicolumn{2}{c}{Unique} & \multicolumn{2}{c}{Multiple} & \multicolumn{2}{c}{Overall} \\
                                &                       & acc@0.25 & acc@0.5 & acc@0.25 & acc@0.5 & acc@0.25 & acc@0.5 \\
        \midrule
        3DVG-Trans \cite{3dvg} & Opt. & 81.9 & 60.6 & 39.3 & 28.4 & 47.6 & 34.7 \\
        3D-SPS \cite{3dsps} & Opt. & \textbf{84.1} & 66.7 & 40.3 & 29.8 & 48.8 & 37.0 \\
        3DJCG \cite{3djcg} & Opt. & 83.5 & 64.3 & 41.4 & 30.8 & 49.6 & 37.3 \\ \midrule
        SAT \cite{sat} & VN & 73.2 & 50.8 & 37.6 & 25.2 & 44.5 & 30.1 \\
        MVT \cite{mvt3d} & PG & 77.7 & 66.5 & 31.9 & 25.3 & 40.8 & 33.3 \\
        ViL3DRel \cite{vil3dref} & PG & 81.6 & 68.6 & 40.3 & 30.7 & 47.9 & 37.7 \\
        \midrule
        \model (scratch) & PG & 76.0 & 66.9 & 33.3 & 27.0 & 41.2 & 34.4 \\
        \model & PG & 77.0 & 67.9 & 37.9 & 30.4 & 45.2 & 37.3 \\
        \model (scratch) & M3D & 77.4 & 70.9 & 38.7 & 34.8 & 45.9 & 41.5 \\
        \model & M3D & 81.6 & \textbf{75.1} & \textbf{43.7} & \textbf{39.1} & \textbf{50.6} & \textbf{45.8} \\
        $\Delta$ & M3D & \inc{4.2} & \inc{4.2} & \inc{5.0} & \inc{4.3} & \inc{4.7} & \inc{4.3} \\
        \bottomrule
    \end{tabular}
\end{table*}

\begin{table}[!h]
\centering
\caption{Captioning results on Scan2Cap dataset. ``C'' stands for ``CIDEr'', ``B-4'' for ``BLEU-4'', ``M'' for ``METEOR'', and ``R'' for ``ROUGE'', respectively. ``@0.25'' and ``@0.5'' represent the overlap ratios between the predicted boxes and ground truth boxes.}\label{tab:scan2cap}
\resizebox{\linewidth}{!}{
\begin{tabular}{lcccc|cccc}
    \toprule
    \multirow{2}{*}{Method} & \multicolumn{4}{c|}{@0.25} & \multicolumn{4}{c}{@0.5} \\
    & C & B-4 & M  & R  & C   & B-4  & M & R \\ \midrule
    Scan2Cap \cite{scan2cap} & 53.7  & 34.3 & 26.1 & 55.0 & 35.2 & 22.4 & 21.4 & 43.5 \\
    3DJCG \cite{3djcg}   & 60.9  & \textbf{39.7} & 27.5 & \textbf{59.0} & 47.7 & 31.5 & 24.3  & 51.8 \\ \midrule
    \model(scratch) & 66.8 & 36.6 & 28.0 & 58.4 & 61.6 & 34.1 & 26.8 & \textbf{55.0} \\ 
    \model & \textbf{71.0} & 36.5 & \textbf{28.4} & 57.6 & \textbf{66.9} & \textbf{34.0} & \textbf{27.1} & 54.3 \\ 
    $\Delta$ & \inc{4.2} & \dec{0.1} & \inc{0.4} & \dec{0.8} & \inc{5.3} & \dec{0.1} & \inc{0.3} & \dec{0.7} \\ 
    \bottomrule
\end{tabular}
}
\end{table}

\section{Experiments}

\subsection{Experimental Settings}
\noindent\textbf{Implementation Details.}  The pre-training runs for 30 epochs with a batch size of 128. We use the AdamW~\cite{loshchilov2019decoupled} optimizer with $\beta_1 = 0.9, \beta_2 = 0.98$.
The learning rate is set to $1e^{-4}$, with a warmup of 3,000 steps, and cosine decay. 
During pre-training, we use ground-truth segmentation masks to generate object-level point clouds.During fine-tuning, we use ground-truth masks or Mask3d~\cite{mask3d}, which depends on the task setting. On the ScanRefer dataset, we also incorporate PointGroup~\cite{pointgroup} for comparison with previous approaches. In ablation studies, we use ground-truth masks in all tasks for simplicity. Both pre-training and fine-tuning are conducted on a single NVIDIA A100 80GB GPU.

\noindent\textbf{3D Visual Grounding.}  We evaluate our model on three datasets for this task: ScanRefer~\cite{scanrefer}, Nr3D, and Sr3D~\cite{referit3d}. For Nr3D/Sr3D, we follow ReferIt3D~\cite{referit3d} to use ground-truth object masks and report the results as the grounding accuracy, \ie, whether the model correctly selects the referred object among ground-truth object proposals. For ScanRefer, we follow \cite{scanrefer} to use detector-generated object proposals and report the results as Acc@$k (k \in \{0.25, 0.5\})$, \ie, the fraction of referring queries whose predicted box overlaps the ground truth with IoU $> k$.

\noindent\textbf{3D Dense Captioning} We evaluate our model on the Scan2cap dataset~\cite{scan2cap} and report the text similarity metrics under different box overlap ratios.

\noindent\textbf{3D Question Answering.}  We evaluate our model on the ScanQA dataset~\cite{scanqa} and use exact matches (EM@1 and EM@10) as the evaluation metric. We also report several sentence evaluation metrics, including BLEU-4, ROUGE, METEOR, and CIDEr. Both test sets (w/ or w/o objects) of ScanQA are used in our evaluation.

\noindent\textbf{3D Situated Reasoning} We evaluate our model on the SQA3D dataset~\cite{sqa3d} and report the answer accuracy under different types of questions as the evaluation metric.


\subsection{Downstream Task Results}
In this section, we discuss the experimental results of the downstream tasks and compare the proposed \model model with the state-of-the-art (SOTA) methods. Results are presented in \cref{tab:nrsr,tab:scanrefer,tab:scan2cap,tab:scanqa,tab:sqa3d,fig:data_efficiency} and the main observations from these results are as follows: 

\begin{table*}[!t]
\centering
\small
\caption{Answer accuracy on ScanQA using object proposals from Mask3D. Each entry  denotes ``test w/ object'' / ``test w/o object''.}\label{tab:scanqa}
\begin{tabular}{lcccccc}
\toprule
Method & \multicolumn{1}{c}{EM@1} & \multicolumn{1}{c}{EM@10} & \multicolumn{1}{c}{BLEU-4} & \multicolumn{1}{c}{ROUGE} & \multicolumn{1}{c}{METEOR} & \multicolumn{1}{c}{CIDEr}\\
\midrule
Image+MCAN \cite{scanqa} & 22.3 / 20.8 & 53.1 / 51.2 & 14.3 / 9.7 & 31.3 / 29.2 & 12.1 / 11.5 & 60.4 / 55.6 \\
ScanRefer+MCAN \cite{scanqa} & 20.6 / 19.0 & 52.4 / 49.7 & 7.5 / 7.8 & 30.7 / 28.6 & 12.0 / 11.4 & 57.4 / 53.4 \\
ScanQA \cite{scanqa} & 23.5 / 20.9 & 56.5 / \textbf{54.1} & 12.0 / 10.8 & 34.3 / 31.1 & 13.6 / 12.6 & 67.3 / 60.2 \\
\midrule
\model (scratch) & 25.2 / 20.4 & 55.2 / 51.5 & 10.5 / 8.7 & 35.5 / 29.6 & 13.8 / 11.6 & 68.6 / 55.7 \\
\model & \textbf{27.0} / \textbf{23.0} & \textbf{57.9} / 53.5 & \textbf{16.0} / \textbf{11.9} & \textbf{38.6} / \textbf{32.8} & \textbf{15.2} / \textbf{12.9} & \textbf{76.6} / \textbf{62.6} \\
$\Delta$ & \inc{1.8} / \inc{2.6} & \inc{2.7} / \inc{2.0} & \inc{5.5} / \inc{3.2} & \inc{3.1} / \inc{3.2} & \inc{1.4} / \inc{1.3} & \inc{8.0} / \inc{6.9} \\
\bottomrule
\end{tabular}
\end{table*}

\begin{table*}[!h]
\begin{minipage}[t]{0.60\textwidth}
\centering
\small
\caption{Answer accuracy on SQA3D using object proposals from Mask3D. Pre-training improves the results of most question types.}\label{tab:sqa3d}
\resizebox{\linewidth}{!}{
\begin{tabular}{lccccccc}
\toprule
\multirow{2}{*}{Method} & \multicolumn{6}{c}{Test set} & \multirow{2}{*}{Avg.} \\ 
\cline{2-7}
& What & Is & How & Can & Which & Other & \\ 
\hline
GPT-3 \cite{sqa3d} & \textbf{39.7} & 46.0 & 40.5 & 45.6 & 36.1 & 38.4 & 41.0 \\ 
ClipBERT \cite{sqa3d} & 30.2 & 60.1 & 38.7 & 63.3 & 42.5 & 42.7 & 43.3 \\
SQA3D(w/o s) \cite{sqa3d} & 28.6 & 65.0 & 47.3 & 66.3 & 43.9 & 42.9 & 45.3 \\
SQA3D \cite{sqa3d} & 31.6 & 63.8 & 46.0 & 69.5 & 43.9 & 45.3 & 46.6 \\
\midrule
\model (scratch) & 32.1 & 62.9 & \textbf{47.7} & 60.7 & 45.9 & \textbf{48.9} & 46.7 \\ 
\model & 34.8 & \textbf{63.3} & 45.4 & \textbf{69.8} & \textbf{47.2} & 48.1 & \textbf{48.5} \\ 
$\Delta$ & \inc{2.7} & \inc{0.4} & \dec{2.3} & \inc{9.1} & \inc{1.3} & \dec{0.8} & \inc{1.8} \\ 
\bottomrule
\end{tabular}

}
\end{minipage}
\hfill
\begin{minipage}[t]{0.38\textwidth}
\centering
\captionof{figure}{\label{fig:data_efficiency} The performance of finetuning \model using various amounts of training data.} 
\includegraphics[width=\linewidth]{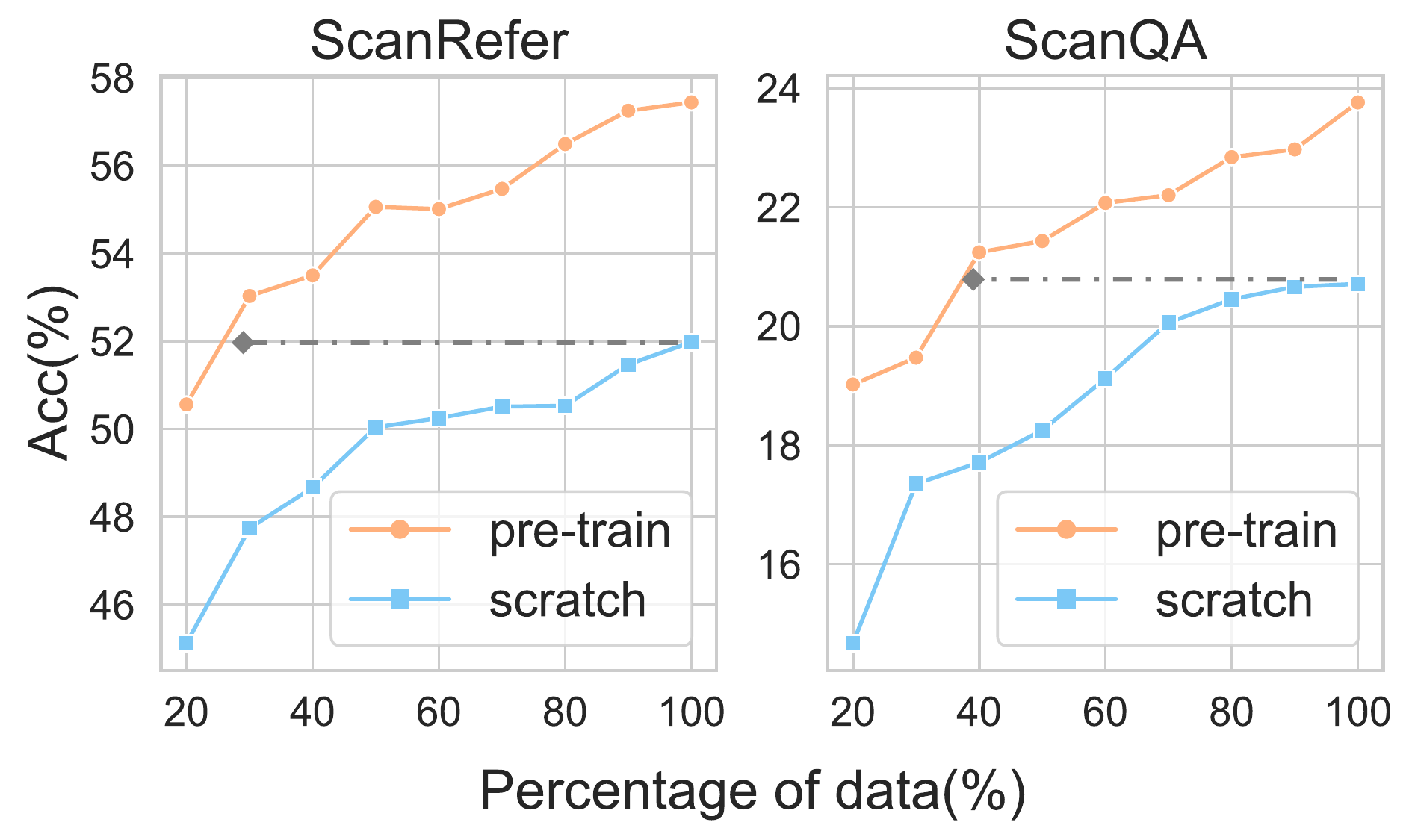}
\end{minipage}
\end{table*}

\begin{enumerate}[leftmargin=*]
    \item \textbf{Even trained from scratch, \model achieves competitive performances with SOTA methods.}
    Specifically, \model (scratch) obtains an overall accuracy of 57.5\% and 69.6\% on Nr3D and Sr3D, which outperforms most previous models; it gets an EM@1 accuracy of 25.2\% on ScanQA, which is 1.7\% higher than SOTA. Of note, \model is trained on these datasets simply using the task losses, without any auxiliary losses or optimization tricks, indicating that \model is a very simple yet effective architecture for 3D-VL tasks.
    \item \textbf{Pre-training on ScanScribe significantly improves the performance of \model.}
    Overall, the pre-training improves the accuracy on Nr3D/Sr3D by 6.7\%/6.8\%, the acc@0.25/0.5 on ScanRefer by 4.7\%/4.3\%, the EM@1 on ScanQA by 1.8\%/2.6\%, the C@0.25 on Scan2Cap by 4.2\%, and the average accuracy on SQA3D by 1.8\%. These large improvements consolidate the efficacy of ScanScribe for the 3D-VL pre-training.
    \item \textbf{The pre-trained \model outperforms SOTA by a large margin.} \model outperforms ViL3DRel~\cite{vil3dref} on Sr3D by 3.6\% and on ScanRefer by 2.7\%/8.1\% (acc@0.25/0.5), beats ScanQA~\cite{scanqa} by 3.5\%/2.1 (EM@1), Scan2Cap SOTA by 10.1\%/19.2\% (C@0.25/0.5), SQA3D~\cite{sqa3d} by 1.9\% (Avg.). \model sets a new record for these 3D-VL tasks and may inspire future research on 3D-VL pre-training.
    \item \textbf{Finetuning \model on downstream tasks with limited annotations achieves strong results.} As shown in \cref{fig:data_efficiency}, being fine-tuned using 30\% and 40\% of the annotations on ScanRefer and ScanQA, the pre-trained \model can achieve better performance than the one trained from scratch with full data. We hypothesize that \model has successfully captured the alignment between 3D objects and text via pre-training and is thus able to readily adapt to downstream tasks of various formats. It also reveals the potential of \model to learn unseen tasks in a zero-shot or few-shot manner, which has emerged in NLP~\cite{gpt-3} and 2D-VL~\cite{flamingo} via large-scale pre-training.
\end{enumerate}

\begin{figure*}[!th]
    \centering
    \includegraphics[width=\linewidth]{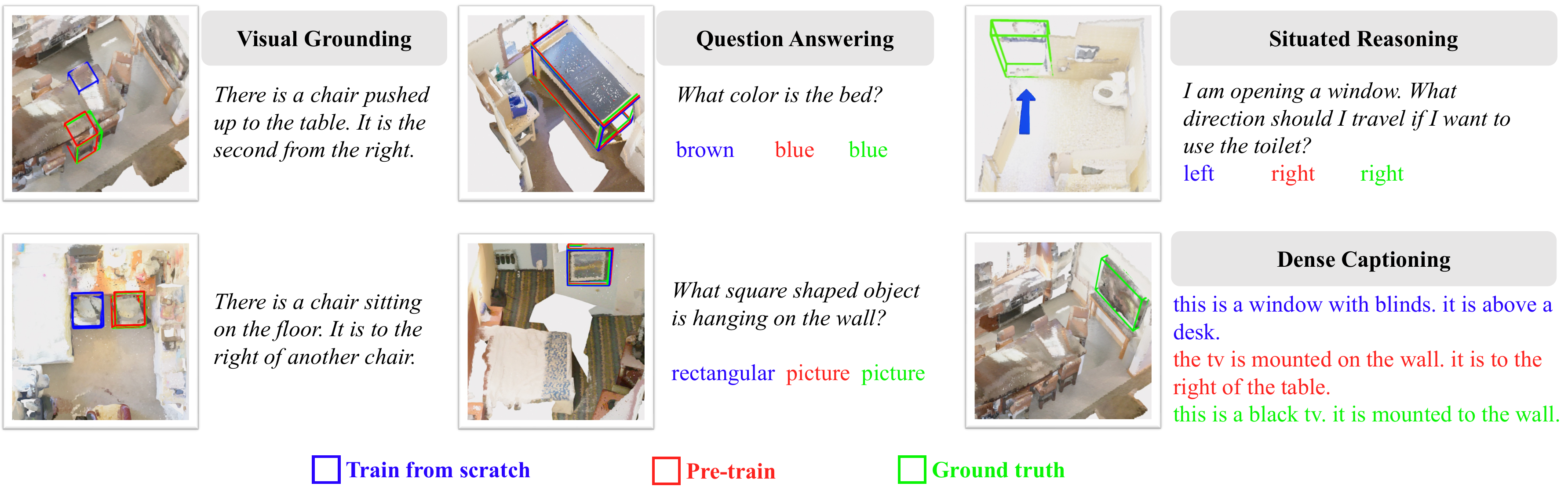}
    \caption{Qualitative results for various tasks. \textit{Italic text} stand for the inputs,  \textcolor{boxblue}{blue boxes or text} for the predictions from \model trained from scratch, \textcolor{boxred}{red} for the predictions from pre-trained \model, and \textcolor{boxgreen}{green} for the ground truth, respectively. The results show that pre-training improves the understanding of spatial relations, visual concepts, and situations.}
    \label{fig:qualititve}
\end{figure*}

\subsection{Ablation Studies}
In this section, we conduct ablation studies to analyze the impact of several important hyperparameters, including Transformer depth, pre-training objectives, and data amount.

\noindent\textbf{Transformer Depth.}
Since the model size is a key factor in the pre-training of NLP and 2D-VL, we study the effect of the transformer depth by varying the number of layers in the multimodal fusion module. As shown in \cref{tab:ablation_layer}, using 4 layers achieves the best performance and simply adding more layers does not help. This observation is somewhat contradictory to the ones from NLP and 2D-VL. It points out that although ScanScribe is much larger than existing 3D-VL datasets, it is still far from enough to unleash the full potential of pre-training in the 3D-VL domain.

\noindent\textbf{Pre-training Objectives.}
\cref{tab:ablation_loss} presents the ablation study for the pre-training objectives. The MLM objective alone slightly benefits question answering (QA), but brings a negative impact on visual grounding (VG). Adding MOM and STM boosts the performance of both QA and VG, which highlights the importance of MOM and STM for aligning 3D vision and text. Overall, using all three objectives together leads to the best performance for both tasks, with STM and MOM providing the greatest improvements in accuracy.

\begin{table}[!ht]
\caption{Ablation studies of \model \wrt Transformer depth, pre-training objectives, and pre-training data. We report the grounding accuracy on ScanRefer for Visual Grounding (VG) and the EM@1 accuracy on ScanQA for Question Answering (QA).}
\label{tab:ablation}
\centering
\begin{subtable}[t]{0.36\linewidth}
    \caption{Transformer Depth}\label{tab:ablation_layer}
    \resizebox{\linewidth}{!}{
    \begin{tabular}{ccc}
    \toprule
    \# layer & VG & QA \\
    \midrule
    2 & 55.8         & 23.7      \\
    4 & 57.4         & 23.8      \\
    6 & 56.6         & 22.8      \\
    8 & 56.3         & 22.7      \\ 
    \bottomrule
    \end{tabular}
    }
\end{subtable}
\hfill
\begin{subtable}[t]{0.62\linewidth}
    \caption{Pre-training Objectives}\label{tab:ablation_loss}
    \resizebox{\linewidth}{!}{
    \begin{tabular}{ccccc}
    \toprule
    MLM            & MOM            & STM          & VG & QA \\ \hline
    $\times$    &  $\times$ & $\times$ & 52.0        & 20.7      \\
    \checkmark    &  $\times$ & $\times$ & 51.5         & 21.3      \\
    \checkmark    &  \checkmark   & $\times$ & 57.1         & 22.5      \\
    \checkmark    &  \checkmark    & \checkmark   & 57.4         & 23.8      \\ 
    \bottomrule
    \end{tabular}
    }
\end{subtable}
\vfill
\begin{subtable}{0.4\textwidth}
    \caption{Pre-training Data}\label{tab:ablation_data}
    \resizebox{\linewidth}{!}{
    \begin{tabular}{ccccc}
    \toprule
    ScanNet    & 3R-Scan     & Objaverse         & VG & QA   \\ \midrule
    $\times$ & $\times$ & $\times$ & 52.0          & 20.7       \\
    \checkmark   & $\times$ & $\times$ & 54.6         & 22.6       \\
    \checkmark  & \checkmark & $\times$  &   56.5        & 23.5       \\
    \checkmark   &  \checkmark & \checkmark  & 57.4          & 23.8        \\ 
    \bottomrule
    \end{tabular}
    }
\end{subtable}
\end{table}

\begin{figure}[!ht]
    \centering
    \includegraphics[width=0.8\linewidth]{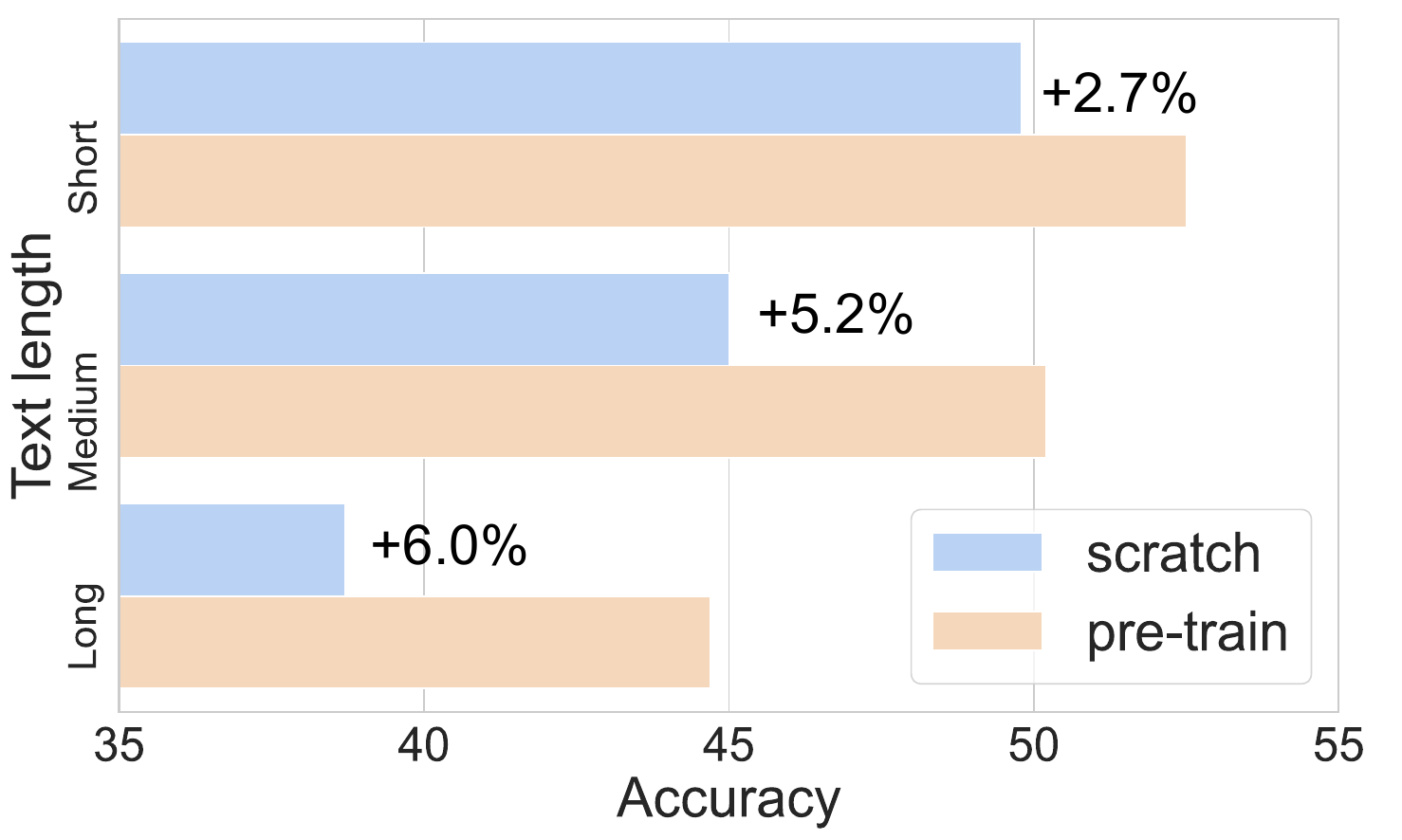}
    \caption{The performance gap between scratch and pre-training over different sentence lengths ($\leq 15, \leq 30, > 30$) in ScanRefer. }
    \label{fig:text_length}
\end{figure}

\noindent\textbf{Pre-training Data.} 
\cref{tab:ablation_data} presents the results using various configurations of pre-training data. We can see that simply using the ScanNet data for pre-training, which is from the same domain as downstream tasks, leads to a significant improvement in VG and QA. This validates the effectiveness of pre-training, even in the case of no additional 3D data than downstream tasks. Adding 3R-Scan and Objaverse increases the amount and the diversity of 3D data, which further boosts the accuracy of both VG and QA. Overall, the best performance for both tasks is achieved when all three data sources are used. This points out a promising path for improving 3D-VL tasks --- collecting more data for pre-training.

\subsection{Qualitative Studies and Additional Results}
In this section, we perform additional studies to better understand how pre-training helps. As shown in \cref{fig:qualititve}, pre-training improves the spatial understanding of \model for visual grounding, so it can better align with human prior viewpoint and reason over spatial relations. This is very helpful when the model needs to distinguish the target object from multiple instances of the same class. Pre-training also helps with a better understanding of visual concepts like colors and shapes, and situations for question answering and situated reasoning. Besides, pre-training enhances the capability of aligning long text with 3D scenes, as evidenced by the larger improvement over longer queries in \cref{fig:text_length}.

%% file: 5.conclusions.tex
\section{Conclusion}
This paper proposes \model, a simple yet effective architecture for 3D-VL tasks. The model simply uses self-attention layers and can be easily adapted to various downstream tasks, without requiring any auxiliary loss or optimization trick. We also introduce ScanScribe, the first large-scale 3D scene-text pairs dataset for 3D-VL pre-training. The pre-trained \model achieves state-of-the-art results on a variety of 3D-VL tasks with superior data efficiency, paving the path to future foundation models for 3D-VL tasks.

\noindent\textbf{Future Works.} Currently, \model uses an offline 3D object detection module, which may be a bottleneck for further improvement. Jointly optimizing the object detection module in the pre-training phase is an interesting future direction. Besides, the data amount in ScanScribe is still insufficient for large-scale 3D-VL pre-training, so scaling up the pre-training dataset as well as the model size is a promising direction to further improve the 3D-VL learning.

%% file: appendix.tex
\appendix
\renewcommand\thefigure{A\arabic{figure}}
\setcounter{figure}{0}
\renewcommand\thetable{A\arabic{table}}
\setcounter{table}{0}
\renewcommand\theequation{A\arabic{equation}}
\setcounter{equation}{0}
\pagenumbering{arabic}
\renewcommand*{\thepage}{A\arabic{page}}
\setcounter{footnote}{0}

{\Large{\noindent\textbf{Appendix}}}
\section{Implementation Details}

\subsection{Downstream Tasks}
\noindent\textbf{ScanRefer~}~\cite{scanrefer}: The ScanRefer dataset contains 51,583 sentences written by humans to describe 800 scenes in ScanNet. We used the official split and allocated 36,665 and 9,508 samples for training and validation, respectively. The dataset is categorized into unique and multiple subsets based on whether the target object is a unique class in the scene. In this task, we need to find the target object described by a sentence.  The evaluation metric for this task is accuracy under intersection over union (IoU) 0.25 and 0.5. 

\noindent\textbf{Nr3D/Sr3D}~\cite{referit3d}: The Sr3D dataset comprises of 83,572 utterances that are automatically generated using a template that focuses on the target-anchor spatial relationship. The Nr3D contains 45,503 human utterances. Both Sr3D and Nr3D are split by ``Easy''/``Hard'' and ``ViewDep''/``ViewIndep''. Hard samples are the ones with two or more distractors in a scene. The view-dependent samples contain language descriptions that rely on viewing directions. These two datasets are also used for visual grounding like ScanRefer. But grounding accuracy with ground truth object proposal is evaluated in this setting.

\noindent\textbf{ScanQA}~\cite{scanqa}: ScanQA is a dataset for 3D question answering with 41,363 questions and 58,191 answers. 
Different from 2D QA, ScanQA focuses more on spatial relations. We follow \cite{scanqa} to use exact matches EM@1 and EM@10 as the evaluation metric. EM@K means the percentage of top K answers from the model matches one of the ground-truth answers. Also, we include text similarity metrics to evaluate answers, including BLEU-4, ROUGE, METEOR, and CIDEr. 

\noindent\textbf{SQA3D}~\cite{sqa3d}: SQA3D is a benchmark for scene understanding of embodied agents with 6.8k unique situations, 20.4k descriptions, and 33.4k diverse reasoning questions. Given a situation, an embodied agent must understand embodied activities, navigation instructions, and common sense, and perform multi-hop reasoning. The evaluation metric is answer accuracy under different types of questions. 

\noindent\textbf{Scan2Cap}~\cite{scan2cap}: Scan2Cap is a dataset for 3D dense captioning. Object descriptions are produced from ScanRefer dataset. For each sentence, two special tokens including [SOS] and [EOS] are added. 

\subsection{Model Architecture}
For the scene encoder, we use a three-layer Pointnet++~\cite{pointnet++} with radius 0.2, 0.4, and sample all points to aggregate a 768-dimension feature. For all text and object tokens, the dimension is 768 in the following multi-modal fusion layers. In the unified encoder, the number of attention heads is set to 12 and the dimension of feedforward layers is set to 2048. For the visual grounding head, we use a two-layer MLP with a hidden dimension of 384. For the question-answering head and the situated reasoning head, we use a two-layer MLP with input dimensions 512 (from the attention flat layer) and 768. 

\subsection{Training settings}
The settings of pre-training including mask ratio, and optimization hyperparameters are introduced in the main paper. We exclude the ScanNet validation and test scenes from pre-training to ensure a fair comparison with other methods. All scenes from 3R-Scan are used for pre-training. In this part, we elaborate on the fine-tuning details.

\noindent\textbf{3D Visual Grounding}: We only use a cross-entropy loss for fine-tuning \model on ScanRefer, Nr3D, and Sr3D. For all these grounding tasks, we set the batch size to 64, and the learning rate to 1e-4,  We multiply the learning rate of the text encoder by 0.1 to stabilize the training process. We fine-tune the pre-trained \model for 100, 100, and 50 epochs for ScanRefer, Nr3D, and Sr3D, respectively. AdamW with $\beta_1=0.9, \beta_2=0.98$ is chosen as the optimizer. We use a warmup of 5,000 steps and a cosine annealing learning rate schedule.  

\noindent\textbf{3D Question Answering}: We use a  cross-entropy answer classification loss and a visual grounding loss for ScanQA. The batch size is 64 and the learning rate is 1e-4. \model is fine-tuned for 30 epochs with 2000 warmup steps for this task. Other optimization parameters are the same as the visual grounding task.

\noindent\textbf{3D Situated Reasoning}:
Answer classification loss is used for SQA3D. We fine-tune \model for 50 epochs. Other optimization parameters are the same as the 3D question-answering task.

\noindent\textbf{3D Dense Captioning}:
Cross entropy loss is used for fine-tuning Scan2Cap. We use the BERT tokenizer to process input sentences and use the casual mask for language transformer. During both fine-tuning and inference, object tokens are not allowed to attend text tokens because of information leaks. \model is fine-tuned for 100 epochs with batch size 64 and learning rate 1e-4. During inference, text tokens are generated by the greedy selection policy.

\subsection{ScanScribe}

\begin{figure*}[!t]
    \centering
    \includegraphics[width=\linewidth]{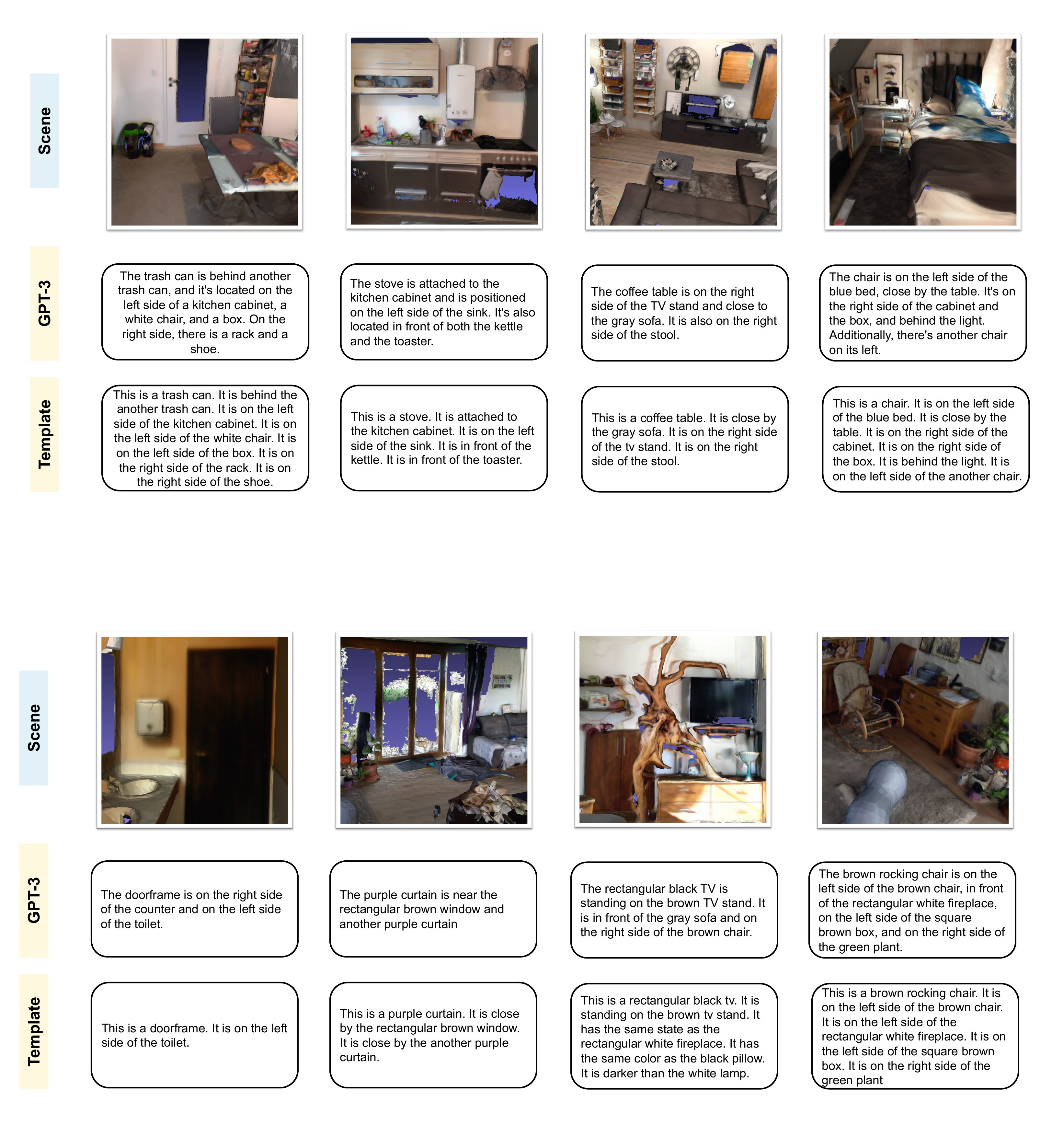}
    \caption{Examples of both template and GPT-3 generated text in ScanScribe dataset. GPT-3 generated text is more natural than template-generated text.}
    \label{fig:language-examples}
\end{figure*}

In the main paper, we introduce our method of generating new scene-text pairs from scene graphs and large language models. More examples and cases are provided in this section. We support 40 relations and the mapping of relations to descriptions for the template-based generation is shown in \cref{tab_relation_description}.

\begin{table}[ht]
\centering
\small
\caption{The mapping of relations to descriptions.} \label{tab_relation_description}
\begin{tabular}{ll}
\toprule
\textbf{Relation} & \textbf{Description} \\
\midrule
supported by & is supported by the \\
left & is on the left side of the \\
right & is on the right side of the \\
front & is in front of the \\
behind & is behind the \\
close by & is close by the \\
inside & is inside the \\
bigger than & is bigger than the \\
smaller than & is smaller than the \\
higher than & is higher than the \\
lower than & is lower than the \\
same symmetry as & has the same symmetry as the \\
same as & is the same as the \\
attached to & is attached to the \\
standing on & is standing on the \\
lying on & is lying on the \\
hanging on & is hanging on the \\
connected to & is connected to the \\
leaning against & is leaning against the \\
part of & is part of the \\
belonging to & is belonging to the \\
built in & is built in the \\
standing in & is standing in the \\
covers & covers the \\
lying in & is lying in the \\
hanging in & is hanging in the \\
same color & has the same color as the \\
same material & has the same material as the \\
same texture & has the same texture as the \\
same shape & has the same shape as the \\
same state & has the same state as the \\
same object type & has the same object type as the \\
messier than & is messier than the \\
cleaner than & is cleaner than the \\
fuller than & is fuller than the \\
more closed & is more closed to the \\
more open & is more open than the \\
brighter than & is brighter than the \\
darker than & is darker than the \\
more comfortable than & is more comfortable than the \\
\bottomrule
\end{tabular}
\end{table}

With these relations, we can use templates like ``This is a \texttt{object}, a \texttt{neighbor} is \texttt{relation} to \texttt{object}'' and utilize GPT-3 to increase text diversity. During pre-training, to balance the proportion of template and GPT-3 generated texts in the 3R-Scan dataset, we duplicate texts from GPT-3 to 15 times for pre-training. Examples from both template-based generation and GPT-3 are presented in \cref{fig:language-examples}. We can observe that given entities and relations in a scene, GPT-3 can summarize them into a fluent and natural sentence.

\section{Additional Results}
\begin{figure*}[ht]
    \centering
    \includegraphics[width=\linewidth]{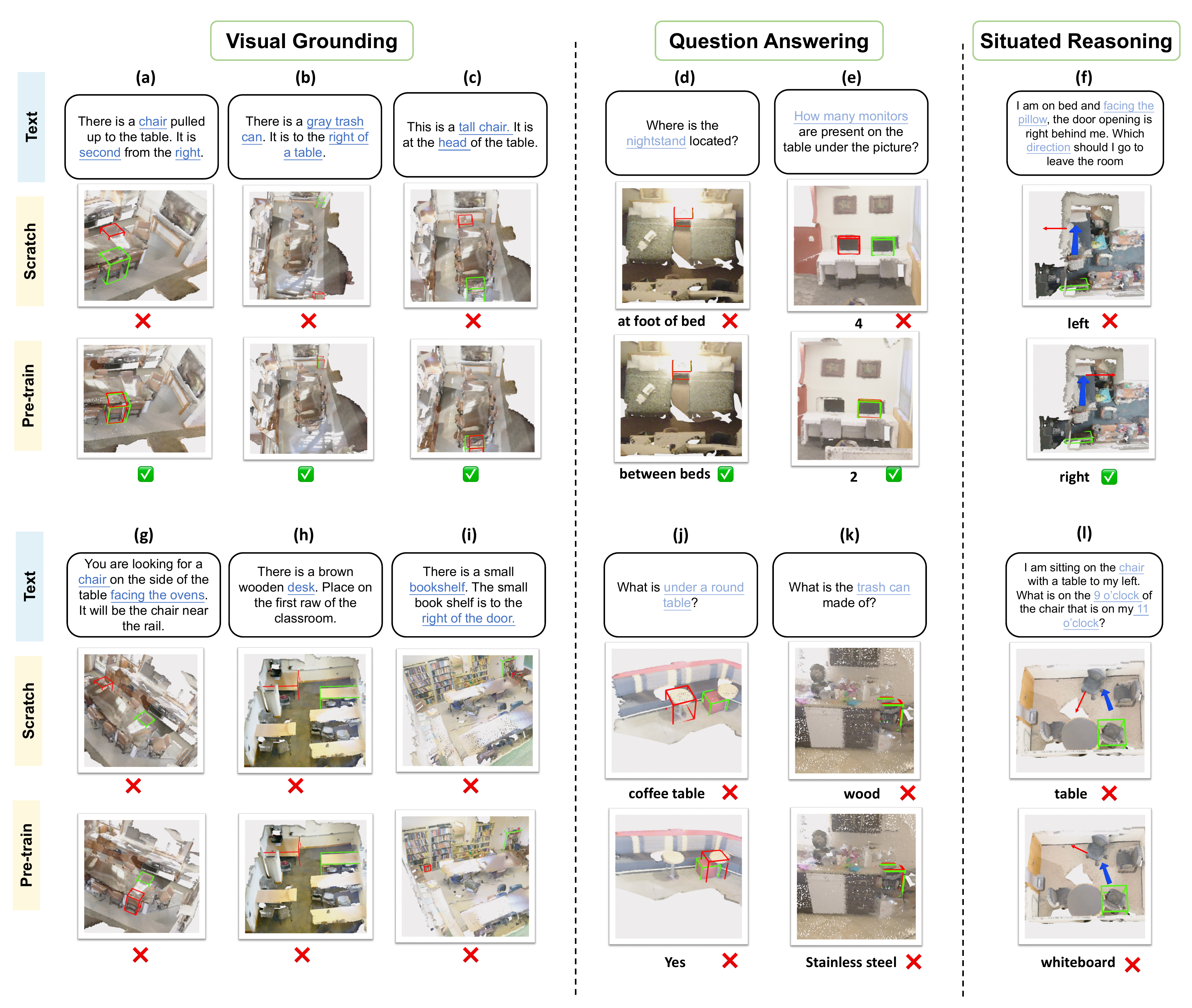}
    \caption{Qualitative results on ScanRefer, ScanQA, and SQA3D. \textcolor{boxgreen}{Green} and \textcolor{boxred}{red} denote the ground-truth and predicted object boxes, respectively. As shown in (a,b,c,d,e,f), the pre-trained \model shows advantages in spatial reasoning, concept grounding, and situation understanding. In spite of these advantages, (g, h, j) indicate that for some complicated cases with spatial relations, the pre-trained model still cannot understand  them. (i, k) show that our model is still limited by the semantic information extracted by point clouds, which fail to locate the right object or understand texture. From (l), we can observe that our model may fail in the case requiring complex multi-hop reasoning.}
    \label{fig:qualititve-sup}
\end{figure*}

We provide ablation studies on the use the template-generated text and GPT-3-generated text. As shown in \cref{tab:ablation_of_human_gpt}, GPT-3-generated text improves Sr3D and Nr3D by 1.0\% and 1.5\%, while having little impact on ScanRefer and ScanQA. More qualitative results including failure cases are provided in \cref{fig:qualititve-sup}.

\begin{table}[ht]
    \centering
    \small
    \caption{Ablation studies on the template and GPT-3 generated text from 3R-Scan. We report the results on ScanRefer, Sr3D, Nr3D and ScanQA.}\label{tab:ablation_of_human_gpt}
    \begin{tabular}{cccccc}
    \toprule
    Template    & GPT-3 & ScanRefer & Sr3D & Nr3D & ScanQA   \\ \midrule
    \checkmark   & $\times$ & 57.4 &  75.4        & 62.7 & 23.7      \\
    \checkmark  & \checkmark & 57.4  & 76.4        & 64.2 & 23.8        \\ 
    \bottomrule
    \end{tabular}
\end{table}

